\documentclass{isprs} 
\usepackage{subfigure}
\usepackage{mathrsfs}
\usepackage{setspace}
\usepackage{geometry} 
\usepackage{epstopdf}
\usepackage{framed,multirow}
\usepackage{latexsym}
\usepackage{lineno}
\usepackage{multirow} 
\usepackage{bm}
\usepackage[labelsep=period]{caption}  
\usepackage[british]{babel} 
\usepackage[hang]{footmisc}
\usepackage{amsmath}
\usepackage{tabularx}
\usepackage{booktabs}
\usepackage{hyperref}
\usepackage{longtable}
\usepackage{array}
\usepackage{afterpage}
\usepackage{pdflscape}
\usepackage{lscape}
\usepackage{rotating}
\usepackage{amssymb}
\usepackage{pifont}
\newcommand{\xmark}{\ding{55}}%

\usepackage[inline]{enumitem}
\usepackage{adjustbox}
\usepackage{todonotes}
\usepackage{xcolor}

\usepackage{stfloats}
\usepackage{lipsum}
\usepackage[binary-units=true]{siunitx}
\usepackage{enumitem}

\usepackage[labelfont=bf]{caption}

\newcommand{\greencheck}{{\color{green}\checkmark}}
\newcommand{\redxmark}{{\color{red}\xmark}}

\usepackage[acronym]{glossaries}
\newacronym{ALS}{ALS}{airborne laser scanning}
\newacronym{MLS}{MLS}{mobile laser scanning}
\newacronym{LoD}{LoD}{level of detail}
\newacronym{OGC}{OGC}{open geospatial consortium}
\newacronym{GML}{GML}{geography markup language}
\newacronym{ASAM}{ASAM}{association for standardization of automation and measuring systems}
\newacronym{TLS}{TLS}{terrestrial laser scanning}
\newacronym{UAV}{UAV}{unmanned aerial vehicle}
\newacronym{HD}{HD}{high definition}
\newacronym{RANSAC}{RANSAC}{random sample consensus}
\newacronym{ROI}{ROI}{region of interest}
\newacronym{DEM}{DEM}{digital elevation model}
\newacronym{ICP}{ICP}{iterative closest point}
\newacronym{NLOS}{NLOS}{non-line-of-sight}
\newacronym{SfM}{SfM}{structure from motion}
\newacronym{FME}{FME}{feature manipulation engine}
\newacronym{OSM}{OSM}{OpenStreetMap} 
\newacronym{RMSE}{RMSE}{root mean square error}
\newacronym{CPT}{CPT}{conditional probability table}
\newacronym{DST}{DST}{Dempster–Shafer theory}
\newacronym{BN}{BayNet}{Bayesian network}
\newacronym{GIS}{GIS}{geographic information systems}
\newacronym{PPD}{PPD}{posterior probability distribution}
\newacronym{CI}{CI}{confidence interval}
\newacronym{CL}{CL}{confidence level}
\newacronym{LiDAR}{LiDAR}{light detection and ranging}
\newacronym{CSG}{CSG}{constructive solid geometry}
\newacronym{BIM}{BIM}{building information modeling}
\newacronym{TUM}{TUM}{the Technical University of Munich}
\newacronym{CRS}{CRS}{coordinate reference system}
\newacronym{IFC}{IFC}{Industry Foundation Classes}
\newacronym{TP}{TP}{terrestrial photogrammetry}

\begin{document}

\title{TUM-FAÇADE: Reviewing and enriching point cloud benchmarks for façade segmentation}

\author{
O. Wysocki, L. Hoegner, U. Stilla
}

\address{
	Photogrammetry and Remote Sensing, TUM School of Engineering and Design, Technical University of Munich (TUM), Munich, Germany  \\ (olaf.wysocki, ludwig.hoegner, stilla)@tum.de\\
}







\abstract{


Point clouds are widely regarded as one of the best dataset types for urban mapping purposes.
Hence, point cloud datasets are commonly investigated as benchmark types for various urban interpretation methods.
Yet, few researchers have addressed the use of point cloud benchmarks for façade segmentation.
Robust façade segmentation is becoming a key factor in various applications ranging from simulating autonomous driving functions to preserving cultural heritage.
In this work, we present a method of enriching existing point cloud datasets with façade-related classes that have been designed to facilitate façade segmentation testing.
We propose how to efficiently extend existing datasets and comprehensively assess their potential for façade segmentation.
We use the method to create the TUM-FAÇADE dataset, which extends the capabilities of TUM-MLS-2016.
Not only can TUM-FAÇADE facilitate the development of point-cloud-based façade segmentation tasks, but our procedure can also be applied to enrich further datasets. 




}

\keywords{Point cloud benchmark, Façade segmentation, Semantic segmentation, Review, TUM-FAÇADE, 3D reconstruction}

\maketitle


\section{Introduction}\label{sec:Introduction}
\sloppy



Buildings are one of the most fundamental elements of a city, which is why digital building reconstruction has become such a pivotal issue for the majority of urban studies.
Every building possesses a number of façades, so digital building reconstruction inevitably involves façade reconstruction, too.
This issue has long been regarded as a challenge within the photogrammetry and computer vision communities \cite{musialski2013survey}.

Although state-of-the-art, semantic 3D building models are widely available, they have generalized extruded façades, chiefly owing to their top-view source datasets ~\cite{HAALA2010570}.
This generalized level has been largely regarded as plausible for various applications \cite{biljeckiApplications3DCity2015}.

However, recent developments have led to growing demand for detailed façade reconstruction in a wide variety of applications, including calculating heating demand \cite{nouvel2013citygml}, preserving cultural heritage \cite{Grilli2019}, assessing flood damage \cite{apel2009flood}, simulating wind flow \cite{montazeri2013cfd}, analysing solar potential \cite{willenborgIntegration2018}, and testing automated driving functions \cite{wysocki2021plastic,schwabRequirementAnalysis3d2019}.

Central factors hampering the development of façade reconstruction methods are a lack of generic, façade-grade datasets and shortage of methods that can accommodate a range of architectural façade styles.
While the former can be aided by \gls{MLS} vehicles, which have recently begun delivering dense, street-level point clouds on an unprecedented scale,
the latter requires various benchmark datasets for testing façade segmentation and reconstruction methods.
However, this process is cumbersome and involves costly measurement campaigns as well as laborious, manual work to provide reference objects. 

Despite this, recent years have witnessed a significant growth in urban point cloud benchmark dataset \cite{griffiths_review_2019}, but few of them have addressed the issue of façades segmentation, albeit they frequently include buildings.
\begin{figure}[h!t]
    \centering
    \includegraphics[width=\linewidth]{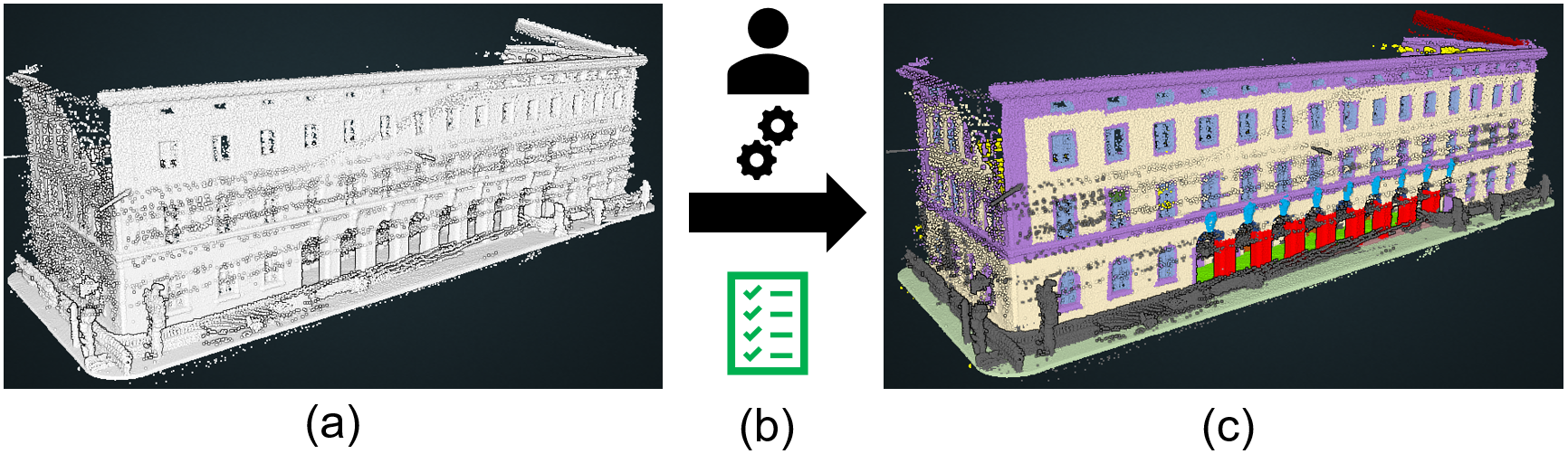}
    \caption{TUM-FAÇADE as a blueprint for enriching existing point cloud benchmarks: a) raw dataset, b) potential assessment, c) extended benchmark by façade classes.}
    \label{fig:bldClasses}
\end{figure}

In this paper, we present a method that reduces the need for creating new point cloud benchmark datasets by enriching existing benchmarks with façade-related semantics.
To this end, our contributions are as follows:

\begin{itemize}
    \item We review the terrestrial, outdoor point cloud benchmark datasets, with a focus on façade segmentation.
    \item We identify terrestrial, outdoor point cloud benchmark datasets that can potentially be used as testing datasets for façade segmentation.
    \item We present a method and classes that can enrich existing point cloud benchmark datasets for façade segmentation methods testing. 
    \item We introduce TUM-FAÇADE\footnote{\url{https://github.com/OloOcki/tum-facade}}  \cite{mediatum1636761}, which enriches the TUM-MLS-2016 \cite{zhu_tum-mls-2016_2020} point cloud benchmark dataset with façade-related classes.
\end{itemize}

\section{Related work}\label{sec:RelatedWorks}


As they have a rich history in the domains of computer vision, photogrammetry, and remote sensing communities, there is a considerable amount of literature on benchmark datasets.
Despite this level of interest, to the best of our knowledge, nobody has ever published a comprehensive review of point cloud benchmark datasets suitable for façade segmentation.

Research has tended to focus on overviews of available point cloud datasets rather than comprehensively reviewing them and focusing on the task of façade segmentation.
For instance, Griffiths and Boehm provide a detailed review of deep learning techniques for 3D datasets, including a chapter concerning benchmark datasets \cite{griffiths_review_2019}. 
Not only do they present benchmark datasets for RGB-D, indoor, and outdoor scenes, but they also provide an overview of selected benchmarks. 
Zhu et al. provide an extensive list of outdoor \gls{MLS} benchmark datasets, as well as presenting the TUM-MLS-2016 benchmark dataset \cite{zhu_tum-mls-2016_2020}. 
Li et al. identify such characteristics as the format of the datasets or the number of available classes, albeit only with reference to a few selected benchmark datasets \cite{li2020deep}. 
The work of Matrone et al.~elaborates on the lack of 3D heritage datasets and bridges this gap by introducing the ArCH dataset \cite{matrone2020comparing}.

Façade segmentation methods have been widely studied \cite{musialski2013survey}.
Much is known about methods using images \cite{teboul2012parsing,mathias2016atlas,muller2007image}, largely facilitated by rich façade image datasets benchmarks, such as those by \cite{riemenschneider2012irregular} or \cite{Tylecek13}.
However, as the images are 2D, they have to be processed to facilitate subsequent semantic 3D reconstruction.
On the other hand, 3D point clouds are deemed among the best data sources for urban mapping purposes, as they yield an immediate 3D representation \cite{xu2021towards}.
Of the particular interest are point clouds acquired by \gls{MLS} vehicles thanks to their, high temporal resolution, and the density of the street-level point clouds \cite{wysocki2021plastic}.
This has led to a recent growth in interest in developing methods of parsing façades using point clouds \cite{martinovic20153d,fan2021layout,zolanvari2016slicing}, especially using machine learning methods \cite{matrone2020comparing,liu2020deepfacade}.

However, only a few studies have focused on releasing point cloud façade segmentation benchmark datasets \cite{matrone2020comparing}.
In the literature, there are a few examples of methods that enrich existing datasets by adding new semantic information. 
One of these is the SemanticKITTI benchmark~\cite{semanticKITTI}, which builds upon the KITTI Vision Benchmark \cite{geiger2012cvpr}.
Alternatively, the dataset can be enriched by conducting a repeated measurement campaign to provide another epoch, which is done chiefly for change detection purposes, as in the work by \cite{zhu_tum-mls-2016_2020}.
\begin{table*} 
\captionsetup{size=footnotesize}
\caption{Our proposed classes for point cloud benchmark datasets to facilitate testing of façade segmentation methods.}
\label{tab:identifiedClasses}
\setlength\tabcolsep{5pt} 
\footnotesize\centering

\smallskip 
\begin{tabular}{cccc}
\hline\noalign{\smallskip}
Index&Class&CityGML&Description\\
& & building-related class&\\
\noalign{\smallskip}\hline\noalign{\smallskip} 
1&wall&WallSurface&Walls excluding any decorative elements\\
2&window&Window&Windows excluding any decorative elements\\
3&door&Door&Including garage doors\\
4&balcony&BuildingInstallation&Excluding pillars and other supportive structures\\
5&molding&BuildingInstallation&Decorative static elements adhering to a building (e.g., cornices) \\
6&deco&BuildingInstallation&Decorative elements mounted to a building (e.g., flags, gargoyles, lights)\\
7&column&BuildingInstallation&Excluding cornices (cornice → molding class)\\
8&arch&BuildingInstallation&Only surfaces oriented downwards\\
9&drainpipe&BuildingInstallation&Pipes and rain gutters of a building\\
10&stairs&BuildingInstallation&Stairs excluding support structures (e.g., poles)\\
11&ground surface&GroundSurface&Any other ground surfaces inside a building envelope\\
12&terrain&-&Any other ground surfaces outside a building envelope (e.g., sidewalks)\\
13&roof&RoofSurface&Any surfaces relating to a roof structure (incl. dormers)\\
14&blinds&BuildingInstallation&Window closures open or closed\\
15&outer ceiling surface&OuterCeilingSurface&Ceilings within a building\\
16&interior&-&Measurements that reflect in a building \\
17&other&-&Any other elements\\
\noalign{\smallskip}\hline
\end{tabular}
\end{table*}

To sum up, most of the existing benchmarks were not created for the purpose of façade segmentation.
Moreover, publications either overlook some benchmarks or provide only sparse statistics, which hampers any detailed comparison of their potential for façade segmentation using point clouds.
Although some methods of enriching existing benchmark datasets have been implemented, they are scarce, especially in the filed of façade segmentation.

\section{Developed methodology}\label{sec:Methods}

\subsection{Assessing existing benchmark datasets}\label{sec:reviewMethods}
As the majority of the point cloud benchmarks were not created for the purpose of façade segmentation, the benchmark datasets we consider comply with several requirements: 
They must represent an open-dataset, outdoor scene, depict buildings or at least façades, and consist of point clouds.
We therefore exclude 2D image benchmark datasets such as \cite{Tylecek13,gadde2016learning,riemenschneider2012irregular}.
Furthermore, as a façade represents a front of a building, it implies that indoor-oriented benchmarks, such as \cite{Armeni}, are out of scope, too.
Due to the limited coverage of façade details, we disregard aerial benchmark datasets, such as \cite{varney} as well as automotive datasets that primarily focus on road objects, such as \cite{geiger2012cvpr}. 

We establish that features crucial to the comparison of datasets for façade segmentation tasks should include the following data: 
\begin{enumerate*}[label=(\arabic*)]
\item year,\label{item:dogs}
\item sensor type, \label{item:cats}
\item scalar fields relevant to segmentation (i.e., point position (XYZ), color (RGB), intensity (I), or normals (N)), \item world (i.e., real or synthetic), \label{item:rabbits}
\item total number of points,
\item whether a dataset is georeferenced,
\item in what region it was acquired,
\item number of available classes,
\item whether a building class is available,
\item whether classes relating to façade details are available, and
\item whether the scene is urban or rural.
\end{enumerate*}



\subsection{Creating an extended benchmark dataset}\label{sec:tumFacadeMethods}

We propose 17 classes for façade segmentation, following the approach of Matrone et al., which is based on CityGML, \gls{IFC}, and Art and Architecture Thesaurus (AAT) \cite{matrone2020comparing}.
We increase the number of classes introduced by Matrone et al., while maintaining consistency and backwards compatibility (i.e., it is possible to merge classes for testing on the same datasets).
To facilitate both segmentation and reconstruction tasks, our classes are also consistent with the modeling guidelines for CityGML \gls{LoD}3 building models \cite{grogerOGCCityGeography2012,special_interest_group_3d_modeling_2020}.
We present the classes in \autoref{tab:identifiedClasses}, with their names and indices, a respective building-related CityGML class, and a brief description.


Extending point cloud benchmark datasets by adding new ground-truth classes inevitably necessitates manual work to be performed by trained annotators.
To minimize the effort involved, supporting algorithms can be used to pre-cluster point clouds, as in \cite{zhu_tum-mls-2016_2020}.
In our case, the central aspect is to first cluster objects that belong to the façade and its immediate vicinity and neglect all other objects.
Hence, we propose using point clouds that are georeferenced to clip-out buildings. 
Using the position obtained from the global \gls{CRS}, we obtain point clouds superimposed on \gls{GIS} datasets.
This, in turn, allows us to create buffers around building footprints extracted from vector \gls{GIS} datasets (e.g., CityGML building models or \gls{OSM} buildings).
This ensures to reject a significant proportion of the point clouds and cluster the building-related points per building object, while addressing global point positioning inaccuracies \cite{wysocki2021unlocking}.
Alternatively, when point clouds are not georeferenced or \gls{GIS} datasets are unavailable, existing benchmark points, annotated as buildings, can be used as a pre-cluster for façade-related points.
If the aforementioned cases are not satisfied, the façades must be extracted manually, or else clustering algorithms must be used, similar to \cite{zhu_tum-mls-2016_2020}.

\section{Results}\label{sec:Results}

\subsection{Potential of existing point cloud benchmarks for façade segmentation}\label{sec:reviewResults}

We analyzed 18 point cloud benchmark datasets (i.e., 17 existing ones plus our dataset), the results of which are presented in \autoref{tab:bigTable}.
As expected, most of the datasets were not created for façade segmentation tasks, with only TUM-FAÇADE \cite{mediatum1636761} and ArCH \cite{matrone2020comparing} being designed specifically for this task.
However, the rest of the set revealed significant extension potential for façade segmentation testing. 
\begin{figure}[htb]
    \centering
    \includegraphics[width=\linewidth]{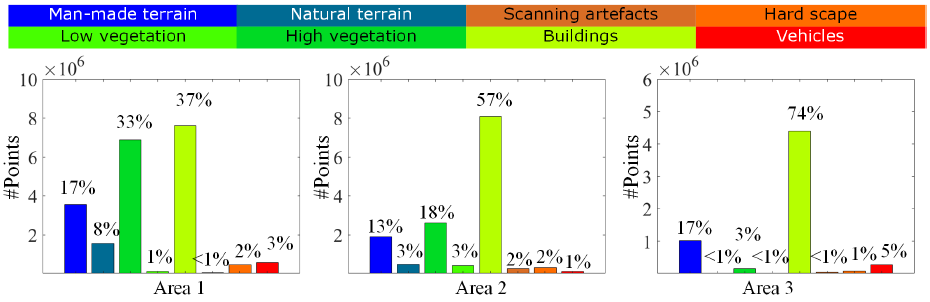}
    \caption{Ratio of annotated semantic classes in the TUM-MLS-2016 benchmark dataset.}
    \label{fig:bldRatio}
\end{figure}

Remarkably, we noticed an increase in the number of datasets released in recent years.
The earliest benchmark in our set is from 2009 \cite{Munoz-2009-10227}, and was the only one to be published that year.
In contrast, in the last three years (i.e., 2020-2022) as many as eight were published, which was almost 50\% of our set.
It is worth noting that no acquisition date is given for the Robotic 3D Scan Repository, either because it is absent in some of the repository's point clouds or varies between them~\cite{robotic3dScan}.

The analyzed list was dominated by \gls{MLS} platforms, 
ranging from an early-type scanner mounted on a car to map a campus in Oakland, Canada \cite{Munoz-2009-10227}, through a backpack mapping unit in the city center of Basel, Switzerland \cite{BimageBlaser}, to a mixture of dense mapping and simulated point clouds in Paris, France \cite{deschaud2021pariscarla3d}. 
Despite this, we included four \gls{TLS} point cloud datasets, as well as the ArCH dataset that combines measurements from \gls{TLS}, \gls{MLS}, an \gls{UAV}, and \gls{TP} measurements \cite{matrone2020comparing}. 

As expected, each point cloud dataset consisted of points with their respective XYZ positions.
However, additional scalar fields varied across the benchmarks.
The intensity values were dominant, being present in 12 datasets of the set.
On the other hand, RGB values occurred in eight datasets, while normals in only two datasets.

Interestingly, our set consisted not only of real-world point clouds but also included synthetic-world point clouds.
This was the case with both SynthCity \cite{griffiths2019synthcity} and Paris-CARLA-3D \cite{deschaud2021pariscarla3d}.
The former presented a completely simulated \gls{MLS} point cloud based on a vector model and covering a combination of European mainland cities and New York, USA~\cite{griffiths2019synthcity}.
The latter combined acquired point clouds with simulated ones using the CARLA environment \cite{deschaud2021pariscarla3d}.

One distinct advantage of synthetic point clouds was that they can easily outnumber the real ones: 700 M to 60 M in the case of Paris-CARLA-3D \cite{deschaud2021pariscarla3d}.
Still, even simulated total points numbers were lower than the ones of \gls{TLS} datasets: the semantic3D.net \gls{TLS} dataset consisted of 4 BN points \cite{hackel2017semantic3d}.
On the other hand, the KITTI-360 \gls{MLS} dataset featured 1 BN points, with 73.7 km of roads being measured in Karlsruhe, Germany \cite{liao2021kitti}.
It was thus clear that quantity of points was directly linked to the pace of acquisition (e.g., \gls{MLS} is intuitively faster than \gls{TLS}) and the total covered area.
The latter is particularly difficult to acquire and compare, since the various datasets have different ways of quantifing this measure, namely: as a number of scenes, the length of road driven, approximate area extent, or else it is unpublished.
Moreover, three obtained datasets \cite{robotic3dScan,ethprs,SydneyDatasetde2013unsupervised} did not reveal their total number of points, thereby this statistic is omitted in these cases.

Regarding point clouds georeferencing, ou analysis showed that the set was equally divided between those published in a local \gls{CRS} and in those in a global \gls{CRS}, both had a score of eight datasets.
It should be noted that semantic3D.net \cite{hackel2017semantic3d}, and ArCH \cite{matrone2020comparing}, provided a description of an acquisition place albeit they were in a local \gls{CRS}. 
Thus, it should be possible to obtain a rough georeference of the point clouds.

Curiously, the majority of the datasets were located in Europe, while two were from North America \cite{Munoz-2009-10227,tan2020toronto}, and there was only one representative both from Asia \cite{dong2020registration} and Australia \cite{SydneyDatasetde2013unsupervised}.
It should be noted that the SynthCity dataset represented a mixture of virtual models from New York, USA, and mainland Europe and thus represented the simulated environment of North America and Europe \cite{griffiths2019synthcity}.  

The most remarkable result to emerge from the analysis is that buildings represent the majority of points in the datasets. 
For example, as we show in \autoref{fig:bldRatio}, ratio of points per building class in the three TUM-MLS-2016 dataset areas outnumbered other classes with scores of 37\%, 57\%, and 74\% \cite{zhu_tum-mls-2016_2020}.
On the other hand, along with the rising classes number, the ratio of points per class vanished.
On average, the number of classes equaled 16.5, with a maximum of 50 and a minimum of 0.
For instance, the Paris-Lille-3D dataset distinguished between 50 different classes, which resulted in classes such as \textit{table} or \textit{mobile scooter}, represented by only 576 and 131 points, respectively \cite{roynard2018parisLille}.
As anticipated, the datasets that focused on the registration of point clouds \cite{dong2020registration,robotic3dScan,BimageBlaser,ethprs} excluded semantic classes.

However, even though the buildings were often annotated, most of the façade-level classes were still absent.
Apart from our TUM-FAÇADE, only the Oakland 3D, Paris-rue-Madame, and ArCH datasets had incorporated façade-level classes in their repositories \cite{Munoz-2009-10227,serna2014parisMadame,matrone2020comparing}.
Yet, although Oakland 3D had a list of several façade-level classes, they were underrepresented; for example, there were 500 and 100 points per stairs and gate classes, respectively \cite{Munoz-2009-10227}. 
The Paris-rue-Madame dataset had a few classes with façades details limited to wall lights, wall signs, and balcony plants \cite{serna2014parisMadame}.
On the other hand, the ArCH dataset, which was designed for façade segmentation purposes, had a rich set of façade-related classes \cite{matrone2020comparing}. 

As we had stipulated that our set had to include point clouds encompassing buildings, most of the datasets' scenes are urban.
Some, such as A2D2 \cite{geyer2020a2d2}, included rural areas, too.
It is worth mentioning that in our set we rejected point clouds from Whu-TLS \cite{dong2020registration}, Robotic 3D Scan Repository~\cite{robotic3dScan}, and ETH PRS \cite{ethprs}, that were indoor or building-unrelated. 

\begin{figure}[htb]
    \centering
    \includegraphics[width=\linewidth]{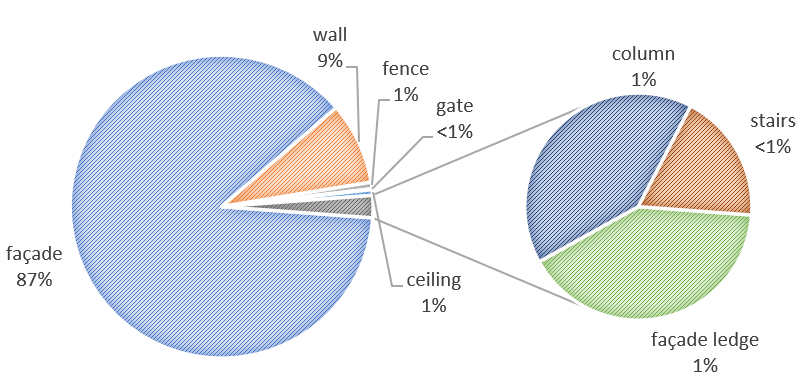}
    \caption{Ratio of annotated façade points per class and low façade classes variability, based on the example of the Oakland 3D point cloud dataset.}
    \label{fig:facadeRatio}
\end{figure}

Moreover, we identified several drawbacks in the currently available point cloud benchmark datasets that hinder effective testing of façade segmentation methods, namely:

\begin{itemize}
    \item \textit{Lack of façade-level classes}: As we present in the \textit{façade-level classes?} column in \autoref{tab:bigTable}, most of the benchmarks do not have any façade-grade classes, which hampers any comparison of methods conducted at such a fine granularity.
    \item \textit{Lack of standardization in façade-level classes}: The classes are inconsistently named and annotated between the datasets and so the meanings of the objects can be confusing, which hinders methods comparison.
    This is exacerbated by the significant variation in the numbers of classes, too, as can be seen in the \textit{\# Classes?} column in \autoref{tab:bigTable}.
    \item \textit{Low variability of façades}: The datasets are limited in a number of façades, with often similar architectural styles. This phenomenon can bias algorithms towards overfitting to a particular architectural style and thus limit their generalization. It can also limit distinction capabilities between important classes such as doors and windows \cite{matrone2020comparing}. For instance, although the TerraMobilita/iQmulus dataset yields high density point clouds, it is limited to a 200 m long survey covering merely a few façades \cite{vallet2015terramobilita}.
    \item \textit{Low ratio of façade-level points per class}: Even when façade-grade classes are available, the number of points per class is low.
    This means that the algorithms can be biased towards highly represented classes (e.g., walls) and neglect the underrepresented ones (e.g., doors).
    We illustrate this drawback in \autoref{fig:facadeRatio}, by analyzing the Oakland 3D point cloud dataset \cite{Munoz-2009-10227}, which is a perfect example of the identified trend.
    \item \textit{Lack of georeferencing}: Many benchmarks do not contain information about the position with reference to the global \gls{CRS};
    They are often provided in a local \gls{CRS}, as shown in \autoref{tab:bigTable}. 
    This excludes or at best hinders a comparison of methods using multimodal sources, such as point clouds in conjunction with 2D or 3D \gls{GIS} datasets, such as in \cite{murtiyoso2019point} or \cite{wysocki2021plastic}.
    \item \textit{Lack of 3D reference building models at LoD3}: Point cloud semantic segmentation algorithms can only be validated against ground-truth labels in point clouds. 
    This means that it is impossible to perform a second-tier validation (e.g., for methods addressing occlusions in point clouds). 
    The application of open semantic volumetric- or surface-based models compliant with at \gls{LoD}3, should enable this process, however. With such models, the benchmarks could be used for both segmentation and 3D reconstruction purposes.
\end{itemize}

\begin{sidewaystable*}
\captionsetup{size=footnotesize}
\caption{Analysis of potential point cloud benchmark datasets for façade segmentation methods testing.} 
\label{tab:bigTable}
\setlength\tabcolsep{5pt} 
\footnotesize\centering
\smallskip 
 \begin{tabular}{cccccccccccc}
        \hline\noalign{\smallskip}
Name  & Year &   Sensor   &   Scalar  &  World   &  \# points   &   Georeferenced?   &   Region   &   \# Classes   &   Building   &   Façade-level   &    Scene\\
   &     &      &   fields   &     &     &     &     &     &   class?   &   classes?   &      \\ \hline
Oakland 3D   &   2009   &   MLS   &   X,Y,Z   &   real   &   1.6 M   &   \redxmark   &   North America   &   44   &   \greencheck   &   $\thicksim$   &   urban\\
ETH PRS   &   2012   &   TLS   &   X,Y,Z,I   &   real   &   \redxmark   &   \redxmark   &   Europe   &   0   &   \redxmark   &   \redxmark   &   urban\\
Sydney Urban Objects Dataset   &   2013   &   MLS   &   X,Y,Z,I   &   real   &   \redxmark   &   \redxmark   &   Australia   &   26   &   \greencheck   &   \redxmark   &   urban\\
 Paris-rue-Madame database   &   2014   &   MLS   &   X,Y,Z,I   &   real   &   20 M   &   \greencheck   &   Europe   &   27   &   \greencheck   &   $\thicksim$   &   urban\\
iQumulus   &   2015   &   MLS   &   X,Y,Z, I   &   real   &   12 M   &   \greencheck   &   Europe   &   8   &   \greencheck   &   \redxmark   &   urban\\
TUM-MLS-2016   &   2016   &   MLS   &   X,Y,Z,I   &   real   &   1.7 BN   &   \greencheck   &   Europe   &   9   &   \greencheck   &   \redxmark   &   urban\\
semantic3D.net   &   2017   &   TLS   &   X,Y,Z, I, RGB   &   real   &   4 BN   &   \redxmark   &   Europe   &   9   &   \greencheck   &   \redxmark   &   rural+urban\\
Paris-Lille-3D   &   2018   &   MLS   &   X,Y,Z,I   &   real   &   143 M   &   \greencheck   &   Europe   &   50   &   \greencheck   &   \redxmark   &   urban\\
SynthCity   &   2019   &   MLS   &   X,Y,Z, RGB,N   &   synthetic   &   368 M   &   \redxmark   &   North America/Europe   &   9   &   \greencheck   &   \redxmark   &   urban\\
A2D2   &   2020   &   MLS   &   X,Y,Z,I   &   real   &   387 M   &   \redxmark   &   Europe   &   38   &   \greencheck   &   \redxmark   &   rural+urban\\
ArCH   &   2020   &   TLS/MLS/UAV/TP   &   X,Y,Z, RGB, N   &   real   &   136 M   &   \redxmark   &   Europe   &   10   &   \greencheck   &   \greencheck   &   urban\\
Toronto-3D   &   2020   &   MLS   &   X,Y,Z,I, RGB   &   real   &   78 M   &   \greencheck   &   North America   &   8   &   \greencheck   &   \redxmark   &   urban\\
Whu-TLS   &   2020   &   TLS   &   X,Y,Z,I, RGB   &   real   &   551 M   &   \redxmark   &   Asia   &   0   &   \redxmark   &   \redxmark   &   rural+urban\\
BIMAGE Datasets   &   2021   &   MLS   &   X,Y,Z, RGB   &   real   &   840 M   &   \greencheck   &   Europe   &   0   &   \redxmark   &   \redxmark   &   urban\\
KITTI-360   &   2021   &   MLS   &   X,Y,Z, RGB   &   real   &   1 BN   &   \greencheck   &   Europe   &   19   &   \greencheck   &   \redxmark   &   urban\\
Paris-CARLA-3D   &   2021   &   MLS   &   X,Y,Z,I,RGB   &   real+synthetic   &   60 + 700 M   &   \greencheck   &   Europe   &   23   &   \greencheck   &   \redxmark   &   urban\\
Robotic 3D Scan Repository   &   \redxmark   &   TLS   &   X,Y,Z,I   &   real   &   \redxmark   &   \redxmark   &   Europe   &   0   &   \redxmark   &   \redxmark   &   rural+urban\\\hline
TUM-FAÇADE   &   2021   &   MLS   &   X,Y,Z   &   real   &   118 M   &   \greencheck   &   Europe   &   17   &   \greencheck   &   \greencheck   &   urban\\

        \noalign{\smallskip}\hline
        \end{tabular}

\end{sidewaystable*}


\subsection{The TUM-FAÇADE benchmark}\label{sec:tumFacade}
In this paper, we present TUM-FAÇADE: a point cloud benchmark dataset that aims to facilitate the development of façade segmentation methods \cite{mediatum1636761}.
Not only does it consists of 17 detailed ground-truth classes but it is based on the challenging \gls{MLS} point cloud dataset, too.
We created TUM-FAÇADE on the basis of the TUM-MLS-2016 benchmark dataset \cite{zhu_tum-mls-2016_2020}, as it featured a challenging, urban environment, with realistic, dense, and georeferenced \gls{MLS} point clouds. 
\begin{figure*}[h!t]
    \centering
    \includegraphics[width=\linewidth]{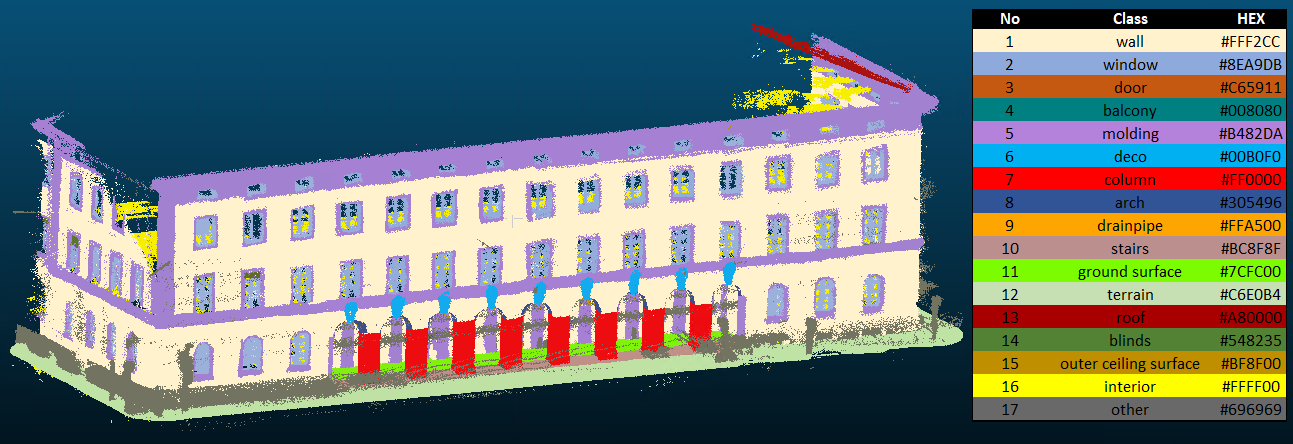}
    \caption{One of the TUM-FAÇADE buildings showing the color-coded points classes.}
    \label{fig:bldClasses}
\end{figure*}
TUM-FAÇADE consists of five annotated and five non-annotated buildings replicating 14 and 15 façades, respectively. There are 17 annotated classes that range from features such as windows to drainpipes, as we show in \autoref{fig:bldClasses} and \autoref{tab:distribution}.
We incorporated local and georeferenced XYZ positions in the scalar fields, together with the respective labels. 
Optionally, the dataset can be enhanced by adding intensity values from TUM-MLS-2016, too. 

To create this dataset, we transformed raw TUM-MLS-2016 point clouds (i.e., 1.7 BN points) to global \gls{CRS} using the transformation matrix included in the TUM-MLS-2016 benchmark repository \cite{zhu_tum-mls-2016_2020}.
Having models aligned in global \gls{CRS} (EPSG: 25832) allowed us to encircle selected building entities with a 3 m buffer, using accurate, cm-grade footprints of governmental CityGML \gls{LoD}2 models\footnote{\url{https://www.ldbv.bayern.de/produkte/3dprodukte/3d.html}}. 
To facilitate the annotation process, each of the selected building point cloud clusters was then shifted to a local \gls{CRS} with an origin in the building's center.

To manually annotate five of these point cloud entities, we used the Semantic Segmentation Editor\footnote{\url{https://github.com/Hitachi-Automotive-And-Industry-Lab/semantic-segmentation-editor}} software by the Hitachi Automotive And Industry Laboratory.
We extended its capabilities to enable it to accommodate our classes, presented in \autoref{tab:identifiedClasses}.
The respective instructions and a configuration file are available under our repository \cite{mediatum1636761}.
We divided the building point clouds into smaller groups of 4 M points to address software, hardware capabilities, and the operator's ability to distinguish between different façade's features. 
Depending on the complexity of an object, labeling took between seven to 23 hours per building, with an estimated total of 83 hours for five buildings or approximately six hours per façade.
\begin{table}[!htb]
\captionsetup{size=footnotesize}
\caption{Annotated classes and points distribution in the TUM-FAÇADE dataset} \label{tab:distribution}
\setlength\tabcolsep{0pt} 
\footnotesize\centering

\smallskip 
\begin{tabular*}{\columnwidth}{@{\extracolsep{\fill}}rcl}
\toprule
  \textbf{\#} & \textbf{Class}  & \textbf{\# points} \\
\midrule
1 & wall & 55,554,783 \\
2 &window & 9,799,964 \\
3 & door & 979,958 \\
4 & balcony & 0 \\
5 & molding&13,497,145 \\
6 &deco&1,104,554 \\
7 &column&1,393,392 \\
8 &arch&220,774 \\
9 &drainpipe&29,398 \\
10 &stairs&419,409 \\
11 &ground surface&7,534,665 \\
12 &terrain&7,918,790 \\
13 &roof&74,035 \\
14 &blinds&547,288 \\
15 &outer ceiling surface&3,797,046 \\
16 &interior&9,477,868 \\
17 &other&5,347,086 \\
  \midrule
  \textbf{Total} &   & 117,696,155 \\
\bottomrule
\end{tabular*}
\end{table}
It should be mentioned that the operator had little prior experience in working with the software.
The second-tier, semi-automatic check was then performed to identify and correct any missing or false annotations.
Once the validity check was completed, the previously used center-shift re-aligned the building to the global \gls{CRS}, and HEX colors were added to the classes as appropiate, as shown in \autoref{fig:bldClasses}.
Another set of five non-annotated buildings for testing, as well as shift and HEX values, are published in our repository \cite{mediatum1636761}.

\section{Conclusions} \label{sec:conclusion}
In this work, we present a comprehensive review of currently available point cloud benchmark datasets with the potential to be used for testing façade segmentation methods. 
We also name potential areas to be addressed in current and future benchmarks. 
To encourage further research and to maximize datasets' potential, we present TUM-FAÇADE, our façade-grade benchmark dataset \cite{mediatum1636761}. 
It enriches the TUM-MLS-2016 benchmark dataset \cite{zhu_tum-mls-2016_2020}, thereby
we show that existing point cloud benchmark datasets can be seamlessly extended by adding façade-grade labels to widen the spectrum of benchmark dataset applications.

We anticipate that the segmentation façade classes we propose, will also facilitate the semantic 3D façade reconstruction process;
the classes are derived from the established CityGML modeling standard \cite{grogerOGCCityGeography2012}.
As such, the classes can be used to identify semantic point cloud clusters and for semantic 3D façade reconstruction.
This facilitates assigning CityGML city model functions, too.
For example, the class \textit{stairs} corresponds to the CityGML class \textit{BuildingInstallation} and function \textit{stairs 1013} \cite{special_interest_group_3d_modeling_2020}.
This enables modeling of CityGML models at \gls{LoD}3 \cite{grogerOGCCityGeography2012}, or at so-called hybrid \gls{LoD} with a façade at \gls{LoD}3 and a roof structure at \gls{LoD}2 \cite{biljecki2016improved}.
Moreover, this feature can also be used for linking the segmented point clouds to existing building models without explicit reconstruction, as demonstrated by \cite{beil2021integration}.

Remarkably, our studies revealed that most of the available point cloud benchmarks not only include a building class but that this class also represents a majority of annotated ground-truth points in the datasets, as we show in \autoref{fig:bldRatio} and \autoref{tab:bigTable}.
Hence, we conclude that most of the existing point cloud benchmarks, although not specifically intended for façade segmentation testing, can be seamlessly extended to serve that purpose.

Moreover, our semantic statistics corroborate that typical \gls{MLS} point clouds can capture fine façade details.
However, they significantly omit roof structures (e.g., only 6\% of points cover roofs in the TUM-FAÇADE benchmark), as we show in \autoref{tab:distribution}.
Thus, as anticipated, \gls{MLS} point clouds are inappropriate for roof segmentation testing purposes.

Nevertheless, we observe that some classes among the various benchmark datasets are inconsistent. 
This hampers the development of generic methods that can be tested on various datasets.
Therefore, to facilitate such developments, we present 17 classes for façade-related annotations.
We believe that they can be used as a set of blueprint classes for further research. 

It remains the case that the outstanding challenge of having overlying ground-truth information of surface- or volumetric-based 3D models with terrestrial point clouds, has not yet been solved \cite{xu2021towards}.
It is worth noting that for several cities and regions city models have been released as open data\footnote{https://github.com/OloOcki/awesome-citygml}.
Yet, they barely overlap with terrestrial point cloud benchmarks, as in the Ingolstadt's \gls{LoD}3 building models\footnote{https://github.com/savenow/lod3-road-space-models} case. 
One of the exceptions is the BIMAGE dataset \cite{BimageBlaser} acquired in Basel, Switzerland, which can be superimposed on open, country-wide, semantic building models.
However, these models are limited in their façades representation, as they consist of \gls{LoD}2 and not \gls{LoD}3 building models.
We believe that this challenge will be the subject of future research.




\section*{Acknowledgments} \label{sec:acknowledgments}
This work was supported by the Bavarian State Ministry for Economic Affairs, Regional Development and Energy within the framework of the IuK Bayern project \textit{MoFa3D - Mobile Erfassung von Fassaden mittels 3D Punktwolken} Grant No.\ IUK643/001.
The work was also carried out within the framework of the \href{https://www.loc.tum.de}{Leonhard Obermeyer Center} at the Technical University of Munich (TUM).
We gratefully acknowledge the team from the Chair of Geoinformatics TUM for their valuable insights and for providing the CityGML datasets.
We are indebted to Jiarui Zhang for his diligent work in the data annotation process.
The authors would like to thank \cite{zhu_tum-mls-2016_2020} for providing us with \autoref{fig:bldRatio}. 

{
	\begin{spacing}{0.7}
		\bibliography{article.bib} 
	\end{spacing}
}

\vspace{1cm}

\end{document}